\documentclass{article}

\usepackage{arxiv}

\usepackage[utf8]{inputenc} 
\usepackage[T1]{fontenc}    
\usepackage{hyperref}       
\usepackage{url}            
\usepackage{booktabs}       
\usepackage{amsfonts}       
\usepackage{nicefrac}       
\usepackage{microtype}      
\usepackage{lipsum}		
\usepackage{graphicx}
\usepackage{natbib}
\usepackage{doi}
\usepackage{verbatim}
\usepackage{subcaption}
\usepackage[misc]{ifsym}

\usepackage{algorithm2e}
\usepackage{amsmath}
\usepackage{array}
\usepackage{multirow}
\usepackage{subcaption}
\usepackage{listings}

\usepackage{pgf}
\usepackage{tikz}
\usetikzlibrary{arrows,automata}
\usetikzlibrary{positioning}

\tikzset{
    state/.style={
           rectangle,
           rounded corners,
           draw=black, very thick,
           minimum height=2em,
           minimum width=2em,
           inner sep=2pt,
           text centered,
           },
}
\usepackage{placeins}
\usepackage{amssymb}
\usepackage{enumitem}
\usepackage{pifont}

\title{Entity Graph Extraction from~Legal Acts -- a~Prototype for a~Use Case in~Policy Design Analysis}

\author{ 
\href{https://orcid.org/0000-0002-3407-7570}{\includegraphics[scale=0.06]{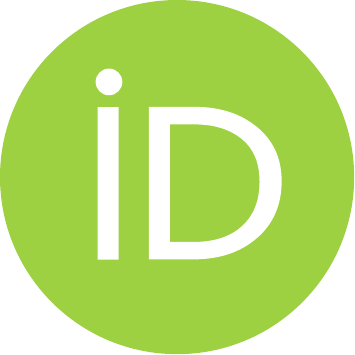}\hspace{1mm}Anna Wr{\'o}blewska$^1$\Letter},
\href{https://orcid.org/0000-0003-2664-2135}{\includegraphics[scale=0.06]{orcid.pdf}\hspace{1mm}Bartosz Pieli{\'n}ski$^2$\Letter},
\href{https://orcid.org/0000-0003-0617-7301}{\includegraphics[scale=0.06]{orcid.pdf}\hspace{1mm}Karolina Seweryn$^1$},
\href{https://orcid.org/0000-0002-8809-5248}{\includegraphics[scale=0.06]{orcid.pdf}\hspace{1mm}Karol Saputa$^1$},\\
 \textbf{Aleksandra Wichrowska}$^1$,
\href{https://orcid.org/0000-0001-5960-8131}{\includegraphics[scale=0.06]{orcid.pdf}\hspace{1mm}\textbf{Sylwia Sysko-Romańczuk}$^3$},
\href{https://orcid.org/0000-0002-6586-8455}{\includegraphics[scale=0.06]{orcid.pdf}\hspace{1mm}\textbf{Hanna Schreiber}$^2$},\\
 	$^1$Faculty of Mathematics and Information Science, Warsaw University of Technology, Warsaw, Poland \\ Email: \texttt{anna.wroblewska1@pw.edu.pl} \\
	$^2$Faculty of Political Science and International Studies, University of Warsaw, Warsaw, Poland \\
 Email: \texttt{b.pielinski@uw.edu.pl} \\
	$^3$Faculty of Management, Warsaw University of Technology, Warsaw, Poland
 }

\date{}

\renewcommand{\shorttitle}{Entity Graph Extraction from  
Legal Acts}

\hypersetup{
pdftitle={Entity Graph Extraction from  
Legal Acts -- a Prototype for a Use Case in Policy Design Analysis},
pdfsubject={cs.LG},
pdfauthor={Anna Wróblewska, Bartosz Pieliński, Karolina Seweryn, Karol Saputa, Aleksandra Wichrowska, Sylwia Sysko-Romańczuk, Hanna Schreiber},
pdfkeywords={Natural Language Processing, Information Extraction, Policy Design, Institutional Grammar, Entity Graph, UNESCO},
}

\begin{document}
\maketitle

\begin{abstract}
This paper presents research on a prototype developed to serve the quantitative study of public policy design. This sub-discipline of political science focuses on identifying actors, relations between them, and tools at their disposal in health, environmental, economic, and other policies. Our system aims to automate the process of gathering legal documents, annotating them with Institutional Grammar, and using hypergraphs to analyse inter-relations between crucial entities. 
Our system is tested against the UNESCO Convention for the Safeguarding of the Intangible Cultural Heritage from 2003, a legal document regulating essential aspects of international relations securing cultural heritage. 
\end{abstract}

\keywords{Natural Language Processing \and Information Extraction \and Policy Design \and Institutional Grammar \and Entity Graph \and UNESCO
}

\section{Introduction}

Recently, we have witnessed a growing interest in policy design analysis~\cite{ResearchHandbookofPolicyDesign,siddiki_2020}. It is focused on examining "the purposeful, functional, and normative qualities of public policies"~\cite{siddiki_2020}. It is primarily concerned with "the content of policies and how this content is organized"~\cite{siddiki_2020,SchneiderIngram}. The interest in policy design can be linked to discussions around specific public policies (e.g., climate policy) and the increasing ease of obtaining large corpora of documents related to these policies. The relatively easy access to large quantities of legal documents and their growing number implies the necessity to analyze them and encourage political scientists to cooperate with computer scientists who have tools for working with large documents. Those tools, mainly scrapers, taggers, and natural language processing (NLP) models, allow retrieving large quantities of relevant documents from the Web, parsing and transforming produced data into easy-to-analyze formats. Computer scientists' research involvement reduces the labour needed for analyzing policy design written into legal documents.

This research and paper concern a prototype built to help develop a specific research area on policy design analysis -- the branch associated with Institutional Grammar (IG). The prototype usage of IG links it to a growing number of studies using IG as a method for retrieving data from written documents. The method allows for analyzing the composition of rules regulating a specific policy in great detail. It identifies links between rules, simplifies their complexity, and identifies vital actors associated with the policy. Although IG is a relatively new concept, it has achieved a high level of standardization~\cite{ig_codebook} which has been acquired by very close cooperation between political and computer scientists; therefore, the prototype, by its design, fits well into the intersection of those two branches of scientific research.     

Graphs are the most popular way of expressing the composition of rules identified by IG. Several published studies have already proved the usefulness of such an approach \cite{Heikkila2018, Olivier-Schlager,MESDAGHI2022120}. It allows looking at elements of rules (e.g., addressees of rules or targets of action regulated by rules) as nodes connected by an edge that is the rule. If an individual rule can be modelled as an edge, then the whole regulation could be seen as a graph or a network. This conceptualization allows introducing measures used in network analysis to study specific policy design, making the entire field of policy design more open to quantitative methods and analysis.

The development of research based on the IG-graph approach meets one significant obstacle: parsing legal text with IG is very labour-intensive. It takes considerable effort to preprocess text for IG analysis and then apply IG itself. Parsing one legal text consisting of over 100 articles can take several weeks. Therefore although the availability of legal documents over the Internet has increased significantly in recent years, up-scaling the IG implementation meets significant obstacles. They have been addressed in two ways: information subtraction and expert moderation. It is possible to focus only on essential -- from the research perspective -- parts of rules embedded in legal texts~\cite{Heikkila2018}. Here, the research focuses on finding only a small set of IG components and performing analysis based only on them. The second approach aims to create systems that parse large quantities of documents in an automated way~\cite{Rice2021}. The prototype of such a system presented in this paper follows the latter path by developing a solution that parses legal texts following IG.

The growing popularity of IG in public policy studies leads to the development of several tools helping to upscale usage of IG. There is the IG Parser created by Christopher Frantz\footnote{\url{http://128.39.143.144:4040/}}, the INA Editor created by Amineh Ghorbani (https://ina-editor.tpm.tudelft.nl/), and attempts have been made to develop a software for automatization of IG annotations ~\cite{Rice2021}. The first tool is handy in the manual annotation of legal texts, and the second is beneficial in semi-manually transforming IG data into a network representation of rules. The third one shows the potential of IG parsing automatization. However, each tool focuses only on one aspect of the multilevel process of transforming a large corpus of legal text into a graph representation. The system presented in this paper chains together many systems that automate the whole process of working with IG, not only its one element. The system is the first step in providing political scientists with a tool that covers the entire spectrum of activities associated with IG implementation in policy design studies. 

Parsing legal texts following IG requires our solution to consist of three technical components: (1) scraper, (2) tagger, and (3) graph modeller. Thanks to that, our prototype can scrap an enormous number of legal documents from the Internet. Then, it can use the IG method, NLP techniques, and IG labelled legal regulations to generate hypergraph representation and analyze inter-relations between crucial entities in policy design. 
The aim is to use IG as an intermediate layer that allows transforming a legal text into a graph which is then analyzed to learn about the characteristic of policy design expressed in the text. A rule, or to use IG terminology, an institutional statement is represented here as an edge connecting actors and objects mentioned in the statement. This approach allows modelling a legal regulation text as a graph and analyzing policy design through indicators used in network analysis. The approach has already been implemented in research papers~\cite{Olivier2019,Olivier-Schlager,MESDAGHI2022120,unpub_Ghorbani2022}.

Noteworthy, the purpose of the designed prototype was not to create an original tool but to build an effective system based on existing technologies. 
All the sub-modules of the system are derived from well-known techniques (scraping, OCR, text parsing, graph building); what is new is their adjusting and coupling. No other system  takes as input regulations distributed around the Internet and produces as an output graph representations of policy design. The biggest challenge for the prototype design was the adjustment of NLP methods for the specificity of IG. Thus, in the following use case, we concentrate on this challenge. We show how to build a graph of inter-relations between important objects and actors based on information extracted with the use of IG from a single but crucial document.


For this paper, our prototype was tested against the UNESCO Convention for the Safeguarding of the Intangible Cultural Heritage (the 2003 UNESCO Convention). This document was chosen because of two main reasons. Firstly, the convention text is relatively short and well written, allowing to test for the first time for the IG tagger to work on a document that does not raise unnecessary confusion both on the syntactic and semantic levels. Secondly, we introduce experts on the 2003 UNESCO Convention to our project, allowing us to check our analysis against their expertise. The 2003 UNESCO Convention is analyzed to compare the position of institutional actors mentioned in the document.  


The following sections present our current achievements leading to our research goal. Section~\ref{sec:assumptions} highlights the current achievements in working with the IG method, which has been developed independently and sketches our current research ideas. Then we describe our prototype architecture and its main modules (Section~\ref{sec:system}). Our use case -- modeling the 2003 UNESCO Convention -- is presented in Section~\ref{sec:use_case}. The paper concludes with a description of our research's upcoming challenges and constraints (Section~\ref{sec:conclusions}).


\section{The Institutional Grammar Layer}
\label{sec:assumptions}

The most crucial part of our system is IG, which is employed as an intermediary layer between the regulation text and its graph representation. IG was used as the layer because it was designed as a schema that standardizes and organizes information on statements coming from a legal text. The statements are understood in IG as bits of institutional information. They have two main functions: to set up prominent institutional actors (organizations, collective bodies, organizational roles, etc.) and to describe what actions those actors are expected, allowed, or forbidden to perform. Therefore IG allows: (1) to identify how many statements are written into a sentence; (2) to categorize those statements; (3) to identify links between them; (4) to identify animated actors and inanimate objects regulated by the statements; and (5) to identify relations between actors and objects defined by the statements.    

\begin{table}[!htb]
 \centering
    \caption{IG main components depending on statement type (regulative or constitutive) based on~\cite[pp.~10-11]{ig_codebook}.\label{tab:ig_components}}
    \begin{tabular}{ |p{2cm}|p{3.5cm}|p{2cm}|p{3.5cm}| }
    \hline
    \textbf{Regulative} & \textbf{Description} &  \textbf{Constitutive} & \textbf{Description}\\
    \hline \hline
    Attribute (A) & The addressee of the statement. &  Constituted Entity (E) & The entity being defined.\\
    \hline
    Aim (I) & The action of the addressee regulated by the statement. & Constitutive Function (F) & A verb used to define (E). \\
    \hline
    Deontic (D) & An operator determining the level of discretion or constraint associated with (I). & Modal (M) & An operator determining the level of necessity and possibility of defining (E).\\
    \hline
    Object (B) & The action receiver described by (I). There are two possible receivers: Direct or Indirect.  & Constituting Properties (P) & The entity against which (E) is defined.\\
    \hline
    Activation Condition (AC) & The setting to which the statement applies. & Activation Condition (AC) & The setting to which the statement applies.\\
    \hline
    Execution Constraint (EC) & Quality of action described by (I) & Execution Constraint (EC) & Quality of (F).\\
    \hline
    \end{tabular}
\end{table}

\begin{figure}[!htb] %
    \centering
    \includegraphics[width=12cm]{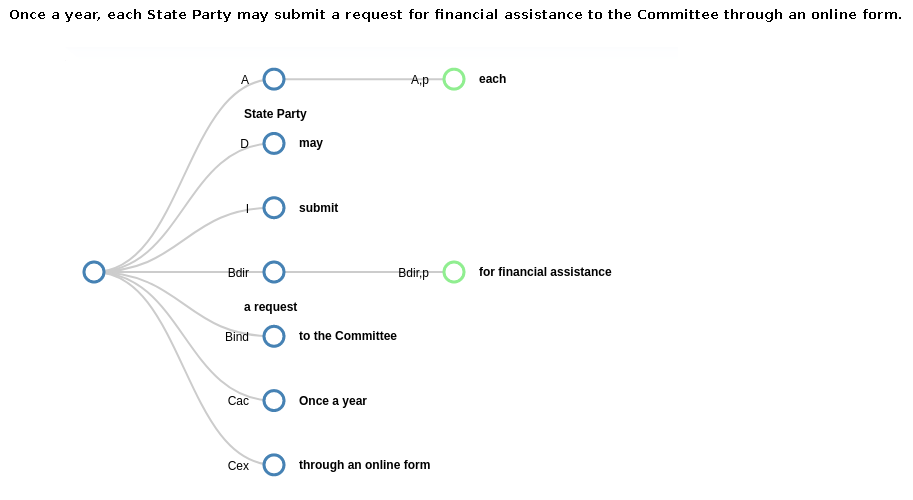}
    \caption{Example of a regulative statement. The tree was generated by the IG Parser https://s.ntnu.no/ig-parser-visual}%
    \label{fig:regulative-example}%
\end{figure}

Figure~\ref{fig:regulative-example} presents an example of a regulative statement based on an alert for the purpose of demonstration sentence from the 2003 UNESCO Convention (Article 23 par. 1). In this statement, "State Party" is an entity whose actions are regulated; therefore, it is annotated as Attribute. An action specifically regulated in the statement is "submit," and this element is identified as Aim. The statement informs us through Deontic "may" that action "submit" is allowed to be performed by State Party. However, the statement also tells through Action Constrain what kind of action of submission is allowed -- "through an online form." We also learn from Activation Condition "once a year" when State Party is entitled to take action "submit." Another piece of information provided by the statement is the kind of target directly affected by the regulated activity -- it is a Direct Object "request" that has the property "for financial assistance." We also learn from the statement which is obliquely affected by State Party action -- it is Indirect Object "the Committee."  

\begin{figure}[!htb] %
    \centering
    \includegraphics[width=12cm]{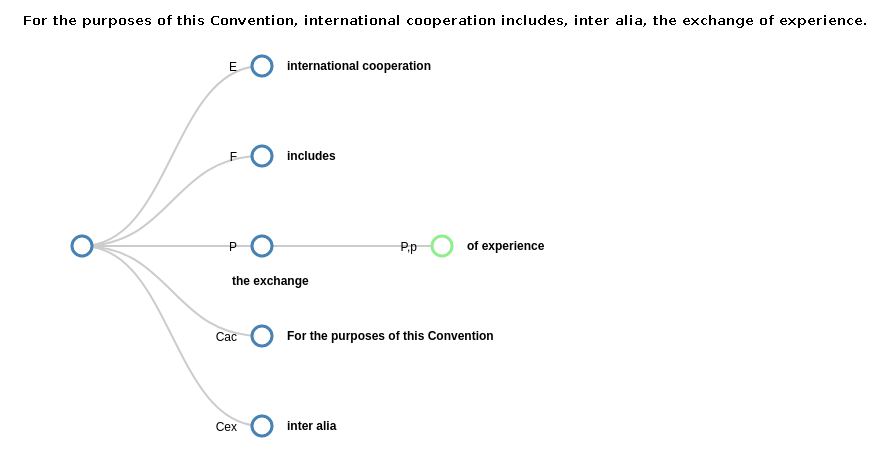}
    \caption{Example of a constitutive statement. The tree was generated by the IG Parser \url{https://s.ntnu.no/ig-parser-visual}}%
    \label{fig:constitutive-example}%
\end{figure}

Another statement, a part of Article 19 par. 1, from the 2003 UNESCO Convention, is a simple example of a constitutive statement (Figure~\ref{fig:constitutive-example}). Here, the statement defines one of the essential properties of "international cooperation" that is indicated as Constituted Entity. The statement informs that international cooperation "includes" (Function) a specific type of exchange (Constituting Property) -- the exchange "of experience" (Constituting Entity property). We also learn that this definition of international cooperation applies only in a specific context -- inside the institutional setting described by the 2003 UNESCO Convention. The information about the context in which the definition applies is provided by Activation Condition: "for the purposes of this Convention."

\subsection{Atomic Statements Extraction}
\label{extraction}

Challenges associated with IG annotation could be divided into three groups: issues with the extraction of atomic statements from sentences, issues with atomic statements classification, and problems with transforming a passive voice into an active one in regulative statements. The challenge raised by \textbf{the extraction of atomic statements} comes from the fact that an atomic statement, as IG defines it, can consist of only single instances of Attribute, Aim, and Object (Direct and Indirect). At the same time, one sentence very often regulates several actions of one actor or even various actors. The way an atomic statement is defined leads to a situation when a relatively simple statement: 
\textit{On the basis of proposals submitted by States Parties, and in accordance with criteria to be defined by the Committee and approved by the General Assembly, the Committee shall periodically select and promote national, subregional and regional programmes, projects and activities for the safeguarding of the heritage which it considers best reflect the principles and objectives of this Convention, taking into account the special needs of developing countries.} It is divided into 18 atomic statements. This fragmentation is caused by the fact that this sentence indicates two Aims -- \textit{select} and \textit{promote} -- and nine Direct Objects:
\begin{itemize}
    \item \textit{national programmes, subregional programmes,} and \textit{regional programmes};
    \item \textit{national projects, subregional projects,} and \textit{regional projects;}
    \item \textit{national activities, subregional activities,} and \textit{regional activities.}
\end{itemize}
At the same time, other IG components are constant:
\begin{itemize}
    \item Attribute: \textit{the Committee};
    \item Deontic: \textit{shall};
    \item Execution Constrains: \textit{"on the basis of proposals submitted by States Parties,"  "in accordance with criteria to be defined by the Committee and approved by the General Assembly," "periodically," and "taking into account the special needs of developing countries."}
\end{itemize}
This configuration of IG components leads to extracting several atomic statements where Aims and Objects are changing, and other components remain the same. All such atomic statements could be expressed in a schematic one: \textit{Based on of proposals submitted by States Parties, and following criteria to be defined by the Committee and approved by the General Assembly, the Committee shall periodically} [Aim] [Direct Object]\textit{, taking into account the unique needs of developing countries.}

These types of challenges are raised even higher by enumerations. The presence of enumeration further increases the number of output atomic sentences. They are usually used in sentences that consist of several compressed ones -- the first part of the sentence is constant when the second part is written as an enumeration element. The challenge here is to identify individual sentences and, secondly, identify separate statements. As a result, Article 13 of the Convention describes State Parties' obligations toward safeguarding intangible cultural heritage that consists of four enumeration elements (where the last element consists of three enumeration elements) at the end is divided into 24 atomic statements. 

\subsection{Classification of Atomic Statements}
\label{classification}

The second challenge associated with IG annotation is \textbf{statement classification}. IG identifies two types of statements: constitutive ones -- definitions and statements describing functions of entities -- and regulative ones -- regulating actions taken by entities. The difficulty of statements classification comes from the similarity of constitutive statements describing the essential role of an entity and regulative statements controlling its behaviour. The institutional statement derived from Article 10 par. 1 -- \textit{The UNESCO Secretariat shall assist the Committee} -- could be understood as a statement regulating the action of the UNESCO Secretariat -- the body is expected to help the Committee. However, the statement is constitutional and assigns a vital function to the UNESCO Secretariat. The UNESCO Secretariat is a large body composed of many sub-groups, each specialising in assisting different UNESCO conventions and programmes. In the case of the 2003 UNESCO Convention, the sub-group in the Secretariat is labelled the Living Heritage Entity (LHE), which is solely dedicated to serving this Convention. The body was created, among other things, to assist the Committee. However, it does not assist any conventional Committee but the specific organ of the 2003 UNESCO Convention: Intergovernmental Committee for the Safeguarding of the Intangible Cultural Heritage. In this case, as in the case of many other statements, proper classification of the statement is impossible without expert knowledge of UNESCO affairs.

\subsection{Passive Voice Transformation}
\label{passive}

The last IG-related challenge regards \textbf{the passive voice}. One of IG's assumptions is that regulative statements can only be expressed as an active voice. In the case of each regulative statement, the Attribute -- the addressee of the statement -- has to be identified. Sometimes, the task is easy. In its original form, Article 1O par. 1 has a form of \textit{The Committee shall be assisted by the UNESCO Secretariat}. All pieces of information needed to transform this sentence into a statement in active voice are provided by the sentence itself. In other cases, an Attribute has to be found in previous sentences or the broader context of the regulation. However, sometimes, it is challenging to determine who should be an Attribute in a statement. Article 30 par. 2 states that \textit{The report shall be brought to the attention of the General Conference of UNESCO}. The sentence does not provide information on who is responsible for bringing attention. The statement's context also does not provide a clear answer to this question -- at least two bodies are mentioned in previous sentences that could be responsible for the action (the Committee and the General Assembly). Only based on the domain expert knowledge, it is possible to determine that the addressee of this statement is the UNESCO Secretariat.

\subsection{Extracting IG Entities from Text} 

Many researchers in the political science domain agree that automated text analysis (e.g., reading, annotating, graph modelling) should become a standard tool for political scientists. However, methodologists and scientists from computer science must contribute new methods, new validation techniques, and adequate technologies with applications~\cite{grimmer_stewart_2013,long2019automatic}.
Many legal and political texts can significantly benefit from the support of automated reasoning. Such support depends on the existence of a logical formalisation of the legal text. Nevertheless, legal documents use a rich language that is not easily accessible to annotate~\cite{Libal_Tomer}.
 
The application of IG allows extracting standardised data on rules from legal texts. However, the IG schema is not associated with any popular way of data analysis, which is challenging to utilise in research. Therefore, researchers have been searching for ways in which IG data could be transformed into more popular data formats. The most promising direction is graph theory. Statements can be seen as edges that connect nodes represented by Attributes, Objects, Constituted Entities, and Constituting Properties. In this regard, a set of edges --statements constitutes a graph representing an individual regulation. This approach allows using popular network analysis indicators to characterise the institutional setting developed by the regulation. A legal act is seen here as a list prescribed by the law relations between different objects and actors. Three approaches have been developed to transform IG data into graph data. The first is represented by T. Heikkila and C. Weible~\cite{Heikkila2018}. IG is used here to identify actors in a variety of legal documents. They are linked in a single document and across many regulations. The aim is to identify a network of influential actors in a particular policy and use network analysis to characterise their associations. Heikkila and Weible used their method to study regulation on shell gas in Colorado and found that the institutional actors governing this industry create a polycentric system. The second approach to IG and network analysis is represented by research by O. Thomas and E. Schlager~\cite{Olivier-Schlager,Olivier2019}. Here, data transformation from IG to graph allows for comparison intensity with which different aspects of institutional actors' cooperation are regulated. The authors used this approach to study water management systems in the USA to link IG data with graph representation. It is built around assumptions that network analysis based on IG allows describing institutional processes written into legal documents in great detail. This approach was also used in a study on the climate adaptation of transport infrastructures in the Port of Rotterdam ~\cite{MESDAGHI2022120}.  

In contrast, our approach discussed in detail in the use case Section~\ref{sec:use_case} is more actor-centred than the previous ones. We aim to compare the positions of prominent actors in a network of rules and use indicators developed by network analysis to describe the place of institutional actors in the formal design of the UNESCO Convention. We also confront these metrics with IG data to determine if an actor's position in a rules network correlates with the place in IG statements.  

Information structured in the form of a graph, with nodes representing entities and edges representing relationships between entities, can be built manually or using automatic information extraction methods. Statistical models can be used to expand and complete a knowledge graph by inferring missing facts~\cite{Nickel2016}.
There are different technologies and applications of language understanding, knowledge acquisition, and intelligent services in the context of graph theory. This domain still is in \emph{statu nascendi} and includes the following key research constructs: graph representation and reasoning; information extraction and graph construction; link data, information/knowledge integration and information/knowledge graph storage management; NLP understanding, semantic computing, and information/knowledge graph mining as well as information/knowledge graph application~\cite{inbook-knowledge-graph}.
Graph theory concepts potentially apply in computer science for many purposes. However, in the last decade, the use of graphs in various fields of social science and human life has become a source of innovative solutions~\cite{graph-theory-2020, Application-Graph-Theory-2021, Andreas-Luschow-2021}.

The collective action approach is the foundation of preparing new scientific research methods and solving current social sciences problems,  which affects the structure of stakeholder networks differently in policy settings. Interactions do not usually occur in an institutional vacuum; they are guided and constrained by agreed-on rules. It is also essential to understand the parameters that drive and constrain them. Hence, the design of institutional behaviours can be measured through Networks of Prescribed Interactions (NPIs), capturing patterns of interactions mandated by formal rules~\cite{tomas-olivier-2019}.
Social network analysis offers considerable potential for understanding relational data, which are the contacts, ties, and connections which relate one agent to another and are not reduced to the properties of the individual agents themselves. So, in network analysis, the relations are treated as expressing the linkages between agents. That analysis consists of a body of qualitative measures of network structure. That relational data are central to the principal concerns of social science tradition, emphasising the investigation of the structure of social action. The structures are built from relations, and the structural concerns can be pursued through the collection and analysis of relational data~\cite{Social-Network-Analysis}.
Our approach uses social network analysis powered by institutional grammar (IG).

\section{Our Prototype}
\label{sec:system}

Our main goal for the prototype is to provide tools that facilitate research in the field of political science and at least partially automate their work with the policy design analysis. 

\subsection{Data Workflow}
\begin{figure}[!htb] %
    \centering
    \includegraphics[width=12cm]{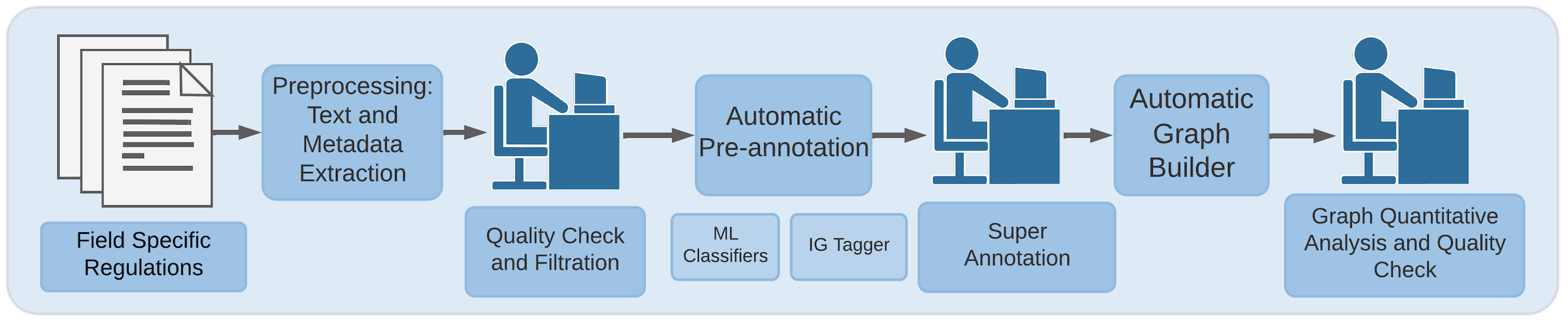}
    \caption{The analysis process of legal acts with our system support and manual work of IG human expert and refinement of a domain expert.}%
    \label{fig:IG_process}%
\end{figure}

The prototype's workflow starts with setting up a crawler used to collect legal documents from websites of interest. Its functionalities mirror the stages of work with legal acts: (1) retrieving documents from different internet resources and (2) selecting, for further analysis, only documents relevant for a specific IG-related research task (filtered by customizable defined rules or keywords). Then, (3) legal texts are prepared, pre-processed, and parsed with IG. The output -- selected IG components are then (4) refined and transformed into a graph where nodes represent institutional actors and objects, and statements containing them are expressed as edges (relations). In this process, we also incorporate quality checks and super-annotations with an IG and domain expert in policy design. Finally, we consider policy design research questions that can be answered using quantitative analysis based on the generated graph. Figure~\ref{fig:IG_process} shows this process of annotating legal documents by incorporating our automated tools in the prototype.
The first significant challenge regarding building the above system is a semantic understanding of legal policy texts. To automate the process of legal text analysis,  we utilize  Institutional Grammar as an intermediate layer for text interpretation and extraction of crucial entities. Thus, in our study, we incorporate IG as a text-semantic bridge. On one side, IG is a tool for describing and analyzing relations between institutional actors in a more systematic and structured way than natural language texts. We used the tagset of IG. However, there are still open issues we currently address with the manual work of policy design researchers. Firstly, there is still a need to define more precisely some IG statements, e.g., context data like Activation Conditions entities. Some entities of IG are not defined precisely and structuralized as a graph structure or formal ontology with classes and logical rules between them. Such structural ordered and precisely defined data (a graph or even an ontology) further can be used, for example, for defining rules of institutions and network analysis and reasoning~\cite{ig_social_network}. Other issues we have yet to address manually are: atomic statement extractions, their transformations into active voice, and the classification of atomic statements -- see Section~\ref{sec:assumptions}.

The second challenge faced by our prototype is to transform IG format with the extracted entities into graph structure and format, which is a proper data structure for automated analysis. Social science has developed, under the label of "social network analysis", significant expertise in expressing information about particular elements of our reality in computer-readable data structures to analyze interrelation between entities. This task is accomplished using IG as an intermediate layer between a legal text and mainly graph-based metrics expressing interrelations between entities. 

\subsection{General Architecture}
\begin{figure}[!htb]
\centering
\includegraphics[,height=0.8\linewidth]{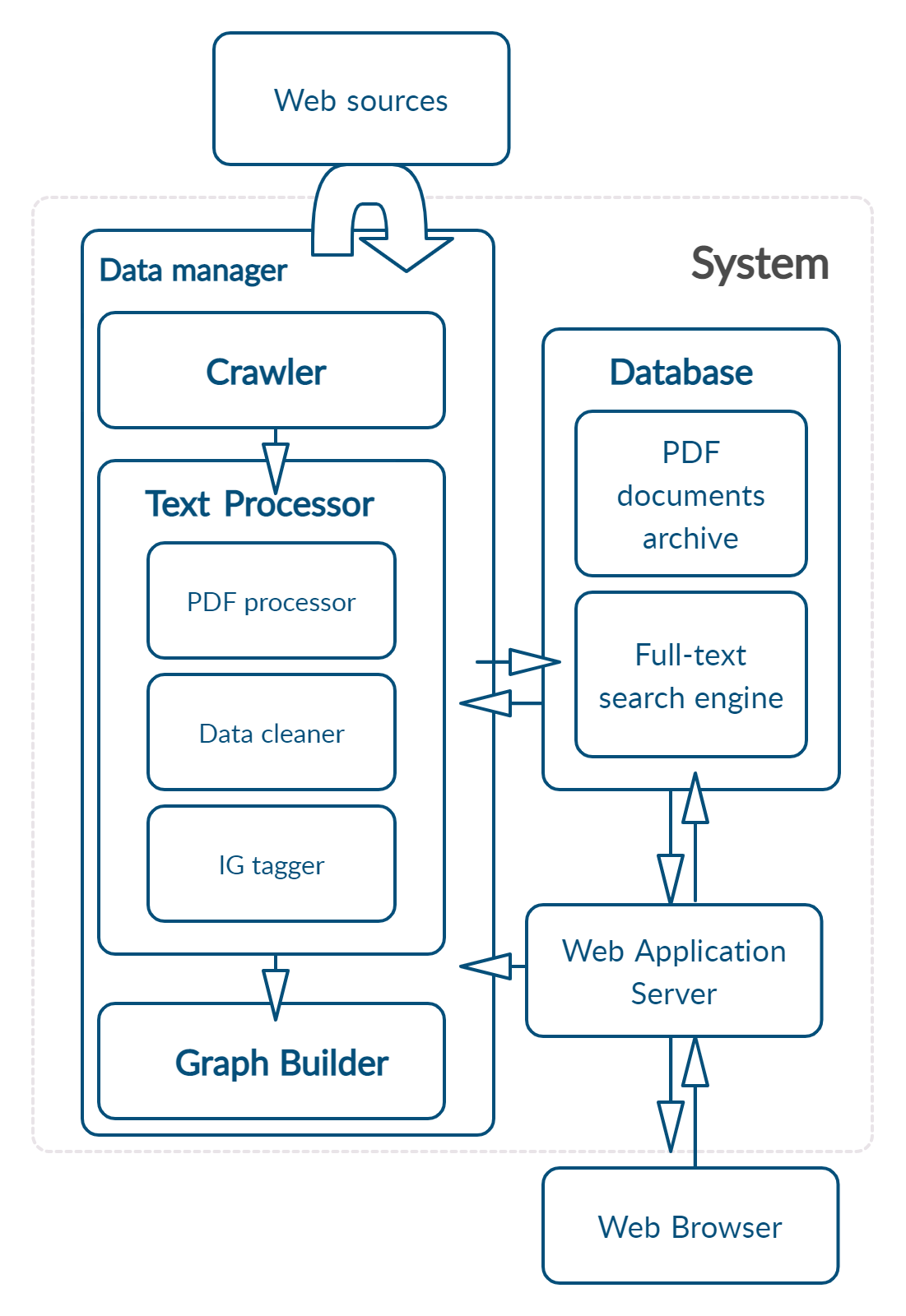}
\caption{Our prototype architecture in a holistic view -- from gathering web documents to browsable results.}
\label{fig:tech_sys_arch}
\end{figure}

Our prototype comprises a set of modules that utilize standard crawlers, taggers, and other NLP models and techniques. These are:
\begin{itemize}
    \item Two databases: for PDF legal act documents and a database for easy access and search through our processed legal acts (with the use of Elasticsearch technology).
    \item Web application server for browsing the documents and using the taggers and other tools.
    \item The core module -- Data manager with four submodules: 
    \begin{itemize}
        \item Crawlers of Internet pages to gain PDF documents with legal acts, 
        \item Text processor for converting PDF documents to extract text and other metadata, and IG tagger with machine learning models to enhance the process of tagging documents.
        \item Rule-based IG tagger, which automates the annotation process.
        \item Graph builder 
        gathering main concepts (institutions) and their relations.
    \end{itemize}
\end{itemize} 
Figure~\ref{fig:tech_sys_arch} presents the general architecture of the prototype.

\subsection{Crawler and Document Databases}


Data acquisition and storage is the first part of the workflow with the legal data. Crawling and databases of crawled data are needed to collect and preserve information for further analysis. In this part, the important prototype modules are (1) crawling, (2) document extractor, (3) data quality filtration, and (4) data warehousing / database (see Figure~\ref{fig:tech_sys_arch}). The technologies used in this stage are standard tool such as most popular scrapers\footnote{\url{https://scrapy.org}}, PDF parsers\footnote{\url{https://github.com/py-pdf/PyPDF2,https://pypi.org/project/pdfminer/}}, and an optical character recognition (OCR) parsers\footnote{Tesseract -- an OCR tool for Python \url{https://pypi.org/project/pytesseract/}.}  for the documents that are not in a text format.

\paragraph{Crawling and Document Processor}
Political scientists need to gather legal acts from chosen institutions and regarding specific topics of interest. As an input to the Crawler tool, we need a list of seed WWW domains for a chosen speciality (a sub-domain for policy design) to be crawled mainly for PDF documents with legal acts. 
Further, the Document Processor extracts and cleans the downloaded PDF documents and their meta-data, such as creation date and keywords, together with URL-related metadata (e.g., country). 

\paragraph{Data Cleaner}
To assure the quality of analysis, the documents must be annotated with the relevance of the subject use case. We provided two criteria for document quality regarding the subject: standard keywords-based and prediction-based. The keywords criterion annotates the document subject by identifying specific keywords related to the use case subject in the text. 

The prediction-based criterion annotates documents with a classifier prediction for being a legal act or not (TF-IDF-based model with 80 features, 96.67 F1-score on a prepared preliminary test set of 102 documents). 
This quality-assessment metadata can be used to filter the data in later steps as search criteria in a full-text search engine.

\begin{figure}[htb] %
    \centering
    \includegraphics[width=12cm]{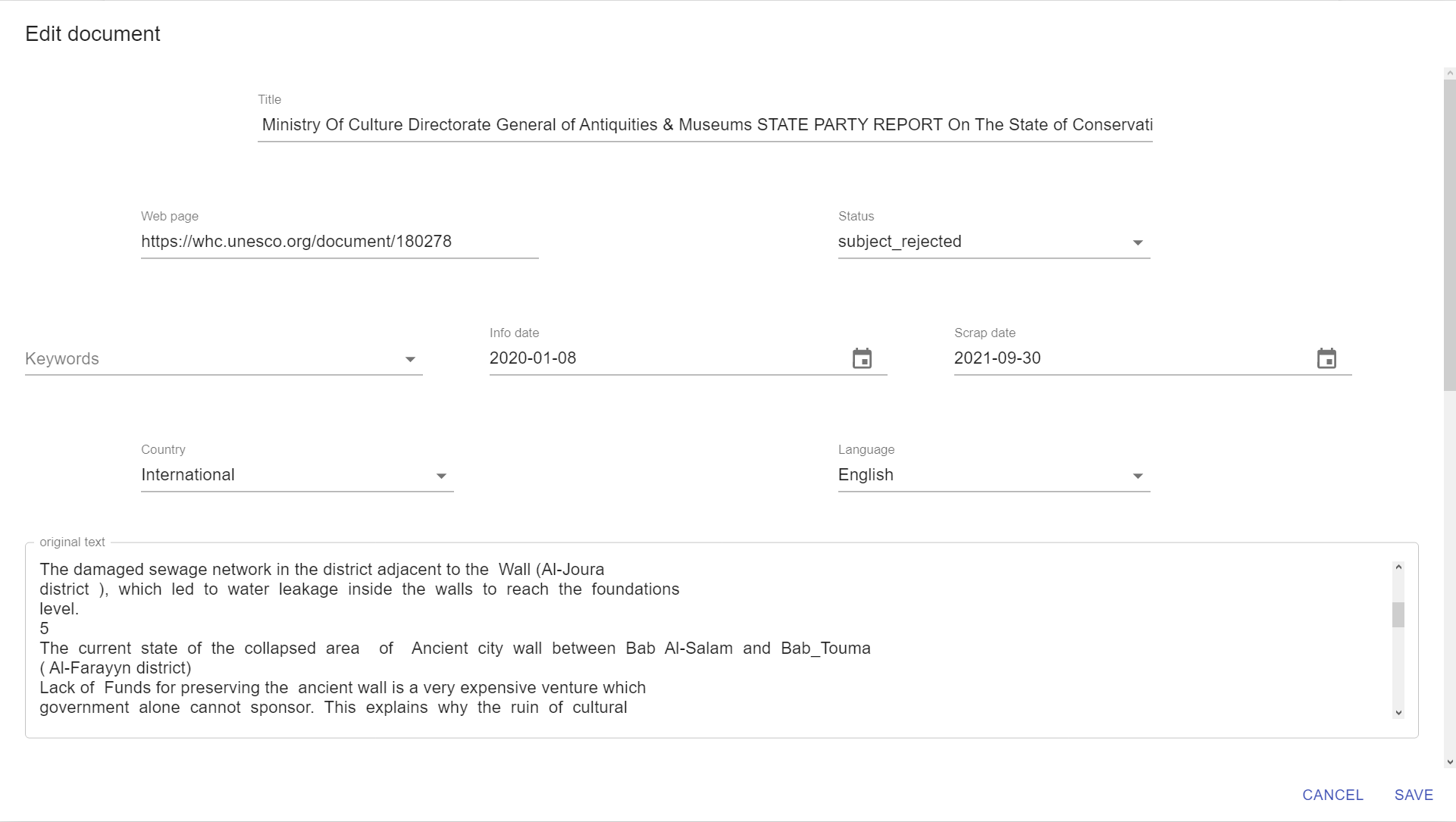}
    \caption{The view of a document in the prototype interface. The user can update fields, select document text fragments to annotate and download or browse the PDF version of a document.}%
    \label{fig:IG_interface_doc}%
\end{figure}

\begin{figure}[htb] %
    \centering
    \includegraphics[width=12cm]{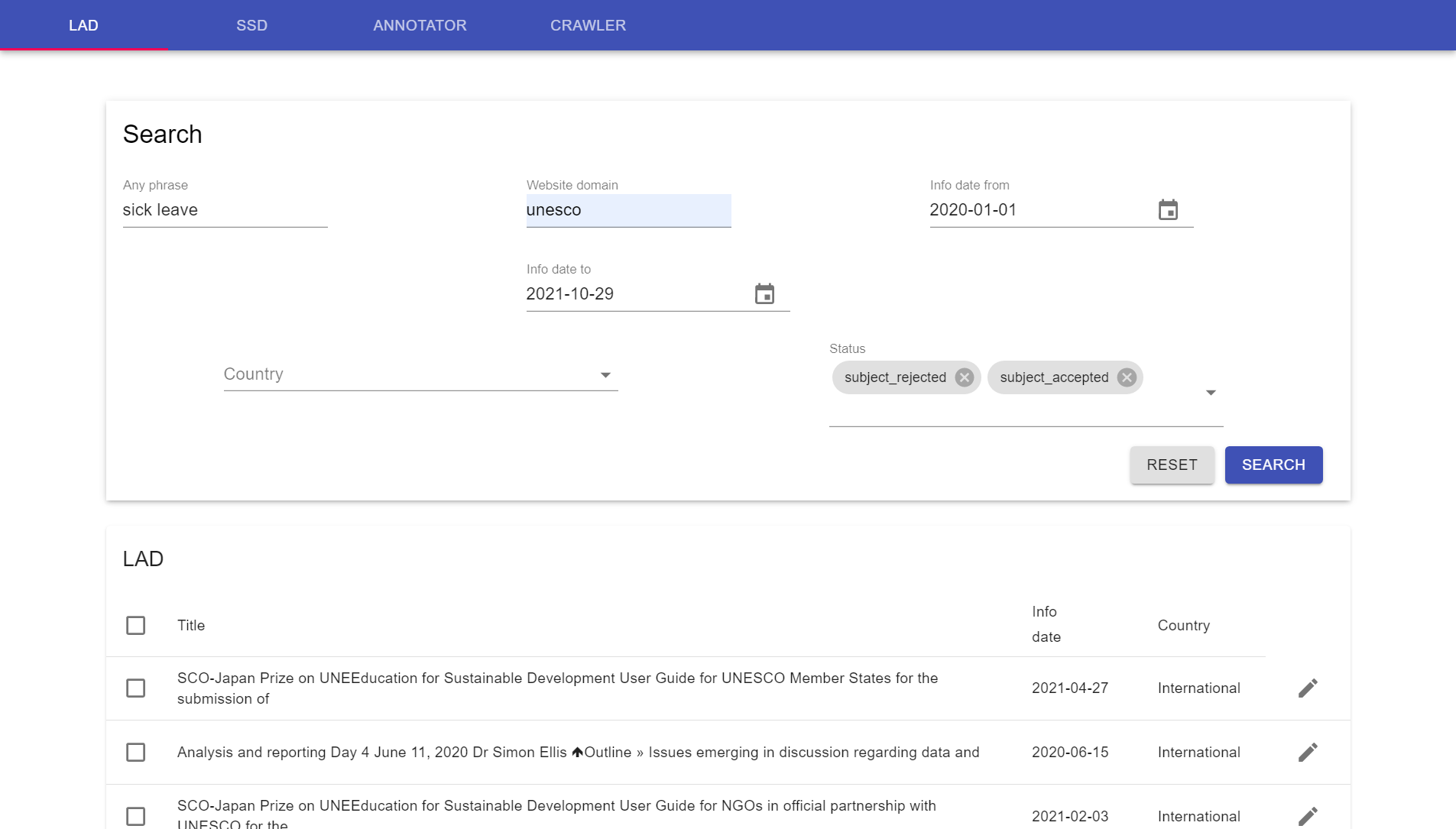}
    \caption{The main view of the prototype interface, which allows searching through a database with different filtering criteria.}%
    \label{fig:IG_interface_search}%
\end{figure}

\paragraph{Database}
The system is intended to provide the gathered data,  e.g. for easily browsing through texts from PDF documents by the domain expert user -- a political scientist. We integrated static PDF file storage and full-text search based on Elasticsearch technology.\footnote{https://www.elastic.co} This engine allows for:  (1) updating documents with new Institutional Grammar annotations and other metadata, (2) searching relevant documents by text content and metadata and filtration of results required for selecting subsets for analysis. Figure~\ref{fig:IG_interface_doc} presents document view with editable details. There is access to metadata, PDF file, and extracted text, as well as the possibility to add IG annotations -- made with our tagger -- to the selected fragment of the document.
Figure~\ref{fig:IG_interface_search} illustrates the main interface window of our prototype, which provides searching capabilities to the database of all crawled documents.

\subsection{IG Tagger}

\paragraph{Statements Pre-processing} 
The institutional statements have two types -- regulative and constitutive -- with a few distinct IG tags. 
The constitutive statements in IG serve for defining purposes, and the regulative statements describe how to regulate behaviours or actions actors can do. For this task -- differentiation of statement types -- we trained a classifier based on TF-IDF with 70 of the most relevant (1,3)-grams and Random Forest to distinguish between these types (train set contained 249 statements). The model's F1-score is 92\% on the test dataset (84 observations out of 382). 
Then, we built the IG tagger consisting of rules specific to regulative and constitutive types of sentences. Similarly, our Graph Builder follows this distinction, adding observations for constitutive and regulative statements because they express additional conditions. 

\paragraph{IG Tagger} 
In the first tagging stage, each word in the statement gets an annotation containing a lemma, a part of speech tag, morphological features, and relation to other words (extracted with Stanza package~\cite{qi2020stanza}). Due to different IG tags in regulative and constitutive statements, the automatic tagger has two different algorithms based on sets of rules dedicated to each type of statement.\footnote{The source code will be available after acceptance.} 

In the following, we show an example of tagger usage. For this purpose, we present only the selected rules used in the analyzed sentence below. 
\begin{enumerate}
    \item \label{item:root_verb} If a sentence contains one word with \textit{root} tag and this word is a verb or an adjective:
    \begin{enumerate}
        \item \label{item:root_verb_function} If the word founded in~\ref{item:root_verb} is a verb, then annotate it as \textit{constitutive function}
        , otherwise as \textit{constituing properties}.
        \item \label{item:root_verb_function_2} If the word annotated in \ref{item:root_verb_function} has a child with \textit{aux:pass} or \textit{cop} relation, then annotate this child as \textit{constitutive function}.
         \item \label{item:entity} If the word annotated in \ref{item:root_verb_function} has a child with one of \textit{nsubj}, \textit{nsubj:pass} or \textit{expl} relation, then annotate this child as \textit{constituted entity}.
        \item \label{item:entity_4} If the word annotated in \ref{item:entity} has a child with one of \textit{det}, \textit{compound}, \textit{mark}, then annotate this child and all child's descendant as \textit{constituted entity}.
        
        \item \label{item:context} If the word annotated in \ref{item:root_verb_function} has a child with one of \textit{obl}, \textit{advmod}, \textit{xcomp}
relation, then annotate this child and all child's descendants as \textit{context}.
    \end{enumerate}
    \item \label{item:modal} If the word with \textit{root} tag has a child with \textit{aux} relation and that child's lemma is one of "must", "should", "may", "might", "can", "could", "need", "ought", "shall", then annotate this word as \textit{modal}.
\end{enumerate}

\begin{figure}[h!]
\begin{center}
\includegraphics[width=10cm]{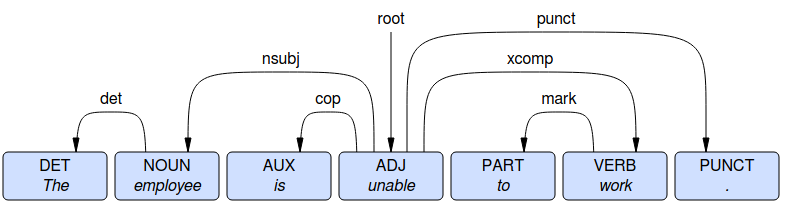}
\end{center}
\caption{Visualization of annotated constitutive sentence by Stanza NLP tagger in CONLL-U Viewer. 
}
\label{fig:conllu_viewer}
\end{figure}

Considering the sentence given in Figure~\ref{fig:conllu_viewer}, our tagging algorithm takes into account the following rules:
\begin{itemize}
    \item According to the rule~\ref{item:root_verb_function} we annotate the word \textit{unable}  as \textit{constituting properties}. 
    \item According to the rule~\ref{item:root_verb_function_2} we annotate the word \textit{is} as \textit{constitutive function}. 
    \item According to the rules~\ref{item:entity} and~\ref{item:entity_4} we annotate words \textit{the employee} as \textit{constituted entity}. 
    \item According to the rule~\ref{item:context} we annotate words \textit{to work} as \textit{context}. 
\end{itemize}

\begin{table}[h!]
    \centering
    \caption{Results of IG Tagger on regulative and constitutive statements.}
    \label{tab:eval_tager_detailed}
    \begin{tabular}{|c|c|c|c|c|}
        \hline 
        \textbf{Layer}  &
        \textbf{Component}  & 
         \textbf{F1 score}  &
         \textbf{Precision} &
         \textbf{Recall} \\
        \hline
        \multirow{6}{*}{\rotatebox[origin=c]{90}{\parbox[c]{1.5cm}{\centering \textbf{Regulative}}}}
        & Attribute &  0.51 & 0.62 & 0.56 \\
        &Object  &  0.55 & 0.67 & 0.61 \\
        & Deontic & 1.0  & 0.96 & 0.98 \\
        &Aim  & 0.99 & 0.86 & 0.92 \\
        & Context &  0.53 & 0.73 & 0.62\\
        &\textbf{Overall} & 0.71 & 0.69 & 0.68 \\
        \hline
        \multirow{6}{*}{\rotatebox[origin=c]{90}{\parbox[c]{1.6cm}{\centering \textbf{Constitutive}}}}
        & Entity & 0.79 &0.64 &0.71  \\
        &Property &0.57  &0.78 &0.66 \\
        &Function & 0.82 &0.82 & 0.82\\
        &Modal &1.00 & 0.83& 0.90\\
        &Context & 0.09& 0.02& 0.03 \\
        &\textbf{Overall} & 0.58 & 0.57 & 0.56 \\
        \hline
    \end{tabular}
\end{table}

Table~\ref{tab:eval_tager_detailed} shows overall tagger performance tested on the 2003 UNESCO Convention consisting of 142 regulative and 240 constitutive statements. For this analysis, predicted and correct tags were mapped before the evaluation: (A, prop) to (A), (B, prop) to (B) for regulative ones, and (E, prop) to (E), (P, prop) to (P) for constitutive statements (prop means property). Applied measures were determined based on the accuracy of the classification of the individual words in a sentence. 
The best results are achieved in recognizing Aim, Deontic, Function and Modal components -- over $99\%$ of F1-score. These tags are precisely defined. As we can see in Table~\ref{tab:eval_tager_detailed}, the constitutive statements and some tags of regulative statements are still a big challenge that should be defined more precisely and solved with machine learning-based text modelling.

\subsection{Graph of Entities Extraction}

We extracted hypergraph information from our analyzed documents to get insights into a given document. The hypergraph vertices are the essential entities -- actors (e.g. institutions) and objects (e.g. legal documents). The statements (in the analyzed documents) represent connections (edges) in this structure. 

Let $H=(V, E)$ be a hypergraph, where verticles $V$ represent actors and objects appearing in the document. $E=(e_{1}, ..., e_{k}), e_{i} \subset V \forall_{i \in 1, ..., k}$ means hyperedges that are created where objects appear together in one statement from the analyzed legal documents. We chose hypergraphs, not graphs, because relations are formed by more than one vertex. 

After building the hypergraph, in our analysis, we use different graph measures, e.g., we adopt a hypergraph centrality definition proposed by authors of~\cite{scentrality}. We can also use other graph metrics to help describe information and relations between entities and define the interrelations and expositions of particular actors.

\section{Use Cases and Impact}
\label{sec:use_case}

Our use case is the 2003 UNESCO Convention, which has established UNESCO listing mechanisms for the safeguarding of intangible cultural heritage, including the Representative List of the Intangible Cultural Heritage of Humanity, the List of Intangible Cultural Heritage in Need of Urgent Safeguarding and the Register of Good Practices in Safeguarding of the Intangible Cultural Heritage~\cite{Blake2020, Schreiber2017}. The convention was chosen as an exemplary case for several reasons: (1) it regulates relations between states regarding a politically charged topic of cultural heritage, (2) there is growing interest in the institutional design of the convention~\cite{Deacon2013}, (3) the convention's text is relatively short and well written, which helps test our system in its initial stage, and (4) the convention with other regulations associated with it (Operational Directives) are the focus of a separate research project. The analysis provided by the use case provides very interesting preliminary data for the project. After further development, the system will be fully incorporated into the project.

As was mentioned previously (Section~\ref{sec:assumptions}), we aim to compare the positions of major actors in the network of rules described by the convention. Following other researchers who transformed IG data into network data~\cite{Heikkila2018,Olivier2019,MESDAGHI2022120}, we use network metrics to describe the locations of actors in institutional settings produced by the 2003 UNESCO Convention. What is new in our analysis is that we compare metrics from network analysis with metrics coming from IG data. The comparison is the first step in understanding if the actor's location in a set of statements is a good predicate of their location in the system created by the regulation. 

The interest in actors' positions in an institutional setting created by the convention comes from the International Public Administration (IPA) literature. One of its main focuses is on the role of IPA against states and other institutional actors~\cite{eckhard_international_2016}. Researchers are interested in what degree 'international civil servants' are autonomous in their decisions and to what extent they can shape the agenda of international organizations. Of particular interest is the position of treaty secretaries -- bodies created to manage administrative issues related to a specific treaty~\cite{bauer_bureaucratic_2016,jorgens_exploring_2016}.    

In our use case, we wanted to learn the Secretariat's position in front of other actors described by the convention, both in data coming from IG and the convention's network representation. The position is here understood as a tension between actors' visibility and centrality. Visibility in the context of legal text analysis is measured by the level of directness in which institutional statements regulate the actor's actions. In the context of IG, it means in which institutional statement components is the actor mentioned. Centrality is associated with actors' positions in a network of statements from a regulation. This analysis assumes that an actor can be mentioned in less prominent IG components. However, at the same time, it could have a prominent role in the formal institutional setting created by a legal document.

Based on the above operationalizations, the research question for the exemplary case could be formulated: What is the relation of visibility versus centrality in the case of the Secretariat?   

\subsection{Exploited Measures}

In this use case, we defined two measures -- visibility and centrality. The visibility is based on Institutional Grammar and envisions how entities are exposed in the analyzed legal document. The centrality is based on the built hypergraph and describes the influence and centrality in the legal act formed in the analyzed document.

\subsubsection{Visibility}
\textit{Visibility} of an actor is defined as the level of the \textit{directness} of how their actions are regulated. It can be measured on an ordinal scale built around the actor's place in a statement (see Table~\ref{tab:directness}). 

\begin{table}[h!]
    \centering
    \caption{The scale of actors' visibility in constitutive (CS) and regulative (RS) statements}
    \label{tab:directness}
    \label{tab:visibility1}
    \begin{tabular}{|p{9cm}|p{1.5cm}|}
        \hline 
        \textbf{Actor:}  &
        \textbf{Weight} \\
        \hline        
        As Attribute in RS or Constituted Property in CS   & 6\\
        As Direct Object in RS or Constituting Entity in CS  & 5\\
        As Indirect Object in RS & 4\\
        In properties of Attribute or Constituted Property & 3\\
        In properties of  Direct Object or Constituting Entity & 2\\
        In properties of Indirect Object  & 1 \\
        \hline 
    \end{tabular}
\end{table}


Actors are ranked by their measures of directness. Each actor is assigned a weight depending on the class (see ranking above). 

$$ visibility = \sum_{c \in \{1, ..., C\}} w_{c} * \frac{n_{c}}{N},$$

where

$N$ - number of statements,

$n_{c}$ - occurrence in class $c$,

$w_{c}$ - rank of class $c$.

Visibility shows the actor's place in the statement and where they were placed in the written text. 

\subsubsection{Centrality}

\begin{figure}[h!]
\centering
  \includegraphics[width=0.85\linewidth]{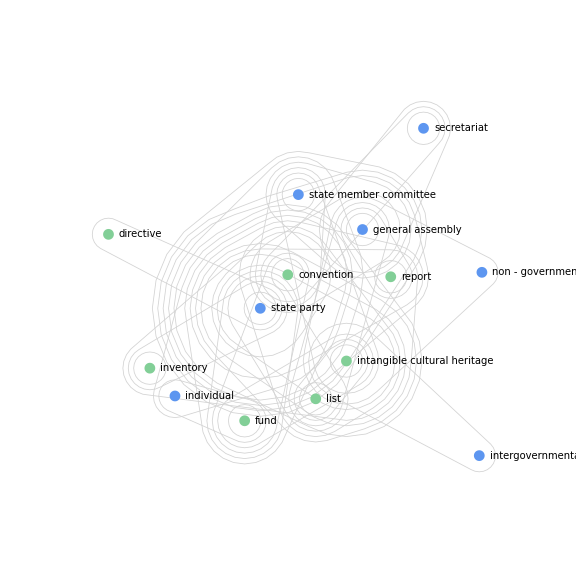}
 \caption{Hypergraph. Blue nodes mean actors, green -- objects.}
\label{fig:graph}
\end{figure}

\textit{Centrality} is not concerned with the actor's placement in any statement but with the actor's placement in institutional design expressed in a regulation. Institutional design is represented here as a hypergraph -- see Figure~\ref{fig:graph}. Each statement forms an edge, and actors are depicted as nodes. 
A set of all edges forms a hypergraph mapping the regulation. The measure of centrality in the hypergraph is computed for each actor. Then, the actors are ranked by their measures of centrality.

\begin{figure}
\centering
\begin{subfigure}{.5\textwidth}
  \centering
  \includegraphics[width=1\linewidth]{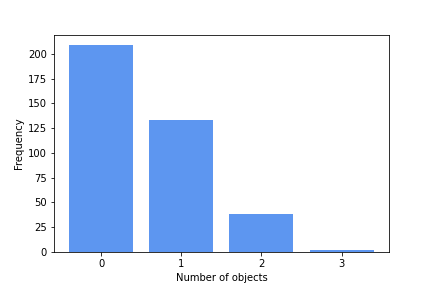}
  \caption{Actors.}
  \label{fig:sub1}
\end{subfigure}%
\begin{subfigure}{.5\textwidth}
  \centering
  \includegraphics[width=1\linewidth]{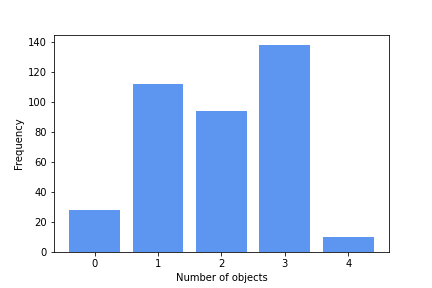}
  \caption{Actors and objects.}
  \label{fig:sub2}
\end{subfigure}
\caption{Histogram of object occurrence in statements which means the number of connected nodes in the hypergraph.}
\label{fig:hist_obj_occurance}
\end{figure}

Figure~\ref{fig:hist_obj_occurance} presents how many actors occur in one statement. The baseline centrality approach analyzes the occurrence of actors together within one sentence. As shown in Figure~\ref{fig:sub1}, usually, only one actor appears within one atomic sentence. Therefore, the analysis was extended, considering the presence of actors and objects such as a report, individual or group. Figure~\ref{fig:sub2} illustrates that this approach has a much greater potential.

\subsection{Use Case Results}

The aim of our use case was, first of all, to investigate potential tensions between locations of actors in institutional statements and their places in a network that models a whole institutional design of the 2003 Convention. Does the placement of an actor in statements predicate its position in the institutional systems? We were particularly interested in the position of the Secretariat. We hypothesized that the Secretariat is a background actor -- it is placed in less important IG components, but it is an essential node in the institutional design of the Convention expressed as a graph. 

\begin{table}
\label{tab:actors_types}
\centering
\caption{Impact of visibility and centrality on actors' relevance in the document.\label{tab:actor_vs_measure}}
\begin{tabular}{|l|l|l|l|} 
\hline
\multicolumn{2}{|l|}{\multirow{2}{*}{}}              & \multicolumn{2}{l|}{\textbf{Centrality}}  \\ 
\cline{3-4}
\multicolumn{2}{|l|}{}                               & \textit{Low}      & \textit{High}         \\ 
\hline
\multirow{2}{*}{\textbf{Visibility}} & \textit{Low}  & minor actor       & background actor~     \\ 
\cline{2-4}
                                     & \textit{High} & overexposed actor & foreground actor      \\
\hline
\end{tabular}
\end{table}

Based on our analysis, we can state that the Secretariat does not hold any particular interesting position. The Secretariat has the lowest level of centrality among all actors and is placed in the middle of the visibility ranking -- see Table~\ref{tab:actor_vs_measure}. Suppose the Secretariat holds a unique position in the 2003 Convention institutional regimes. In that case, it is granted not by the Convention itself but other more operational documents like Directives to the Convention. They should be analyzed in future research. 

\begin{figure}[h!] 
\begin{center}
\includegraphics[width=\textwidth]{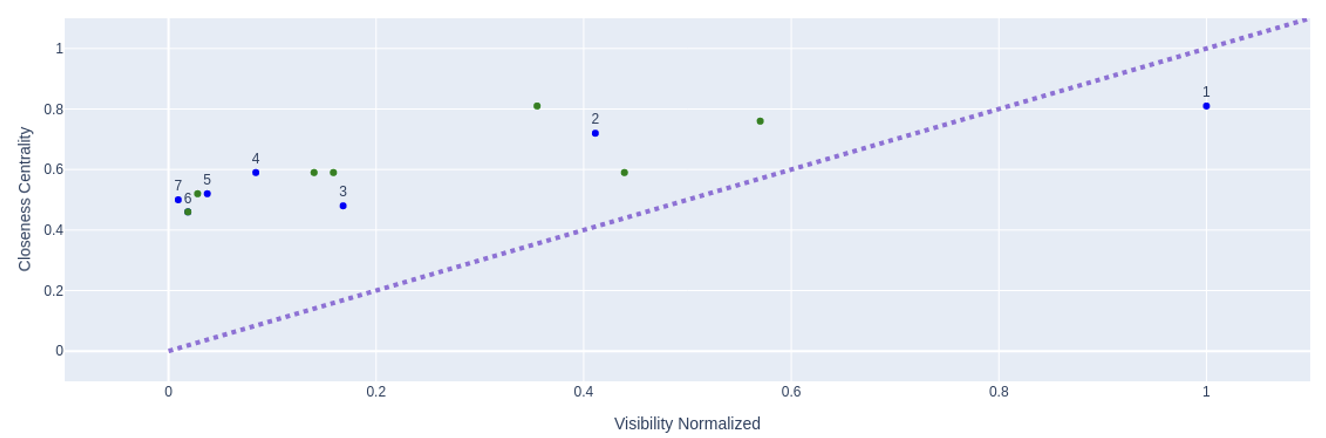}
\end{center}
\caption{Visibility vs Closeness Centrality, where (1) is State Party, (2) is General Assembly, (3) is Secretariat, (4) are States Members of the Committee, (5) is Individual, (6) is Intergovernmental Committee, and (7) is Non-governmental organization. Note: green dots are objects, and blue ones are actors.}
\label{fig:metrics-plot}
\end{figure}

There are, however, two interesting facts that our analysis exposed. The first one is that all actors, except one, have higher levels of centrality than visibility. They are above the dashed line in Figure~\ref{fig:metrics-plot}. They all, including the Secretariat, are "relative" hidden actors -- their scores of visibility scores are significantly lower than their centrality scores. The second interesting fact is that State Party is the only overexposed actor in the 2003 Convention. This actor is placed below the dashed line. However, we can only talk about the relative "over-exposition" of the State Party -- it has, after all, the highest score of centrality among all actors. Based on this observation, we can tell that states that are parties of the Convention are crucial actors in the institutional setting created by the 2003 UNESCO Convention, but not as vital as their place in institutional statements indicates. This observation probably reflects the fact that international conventions are usually formulated in a manner that, above all, regulates the actions of actors being parties of international conventions - states themselves.   

\begin{table}[h!]
    \centering
    \caption{Visibility and centrality of actors appearing in the document.}
    \label{tab:visibility}
    \begin{tabular}{|l|p{3cm}|p{2.8cm}|}
        \hline 
        \textbf{Actor/ object}  &
        \textbf{Visibility} &
        \textbf{Closeness Centrality} \\
        \hline
        State Party & 1.07 & 
        0.81\\
        Convention & 0.61 & 
        0.76\\
        Fund & 0.47 & 
        0.59\\
        General Assembly & 0.44 & 
        0.72 \\
        Intangible cultural heritage & 0.38&  
        0.81\\
        Secretariat  & 0.18 & 
        0.48\\
        List & 0.17 & 
        0.59\\
        Report & 0.15 & 
        0.59\\
        States Members of the Committee  & 0.09 & 
        0.59\\
        Individual & 0.04 & 
        0.52\\
        Inventory & 0.03 & 
        0.52\\
        Intergovernmental Committee & 0.02 & 
        0.46\\
        Directive & 0.02 & 
        0.46\\
        Non - governmental organization &    0.01 & 
        0.5 \\
        Community  & 0 &
        0\\             
        Group & 0 
        & 0\\
        \hline 
    \end{tabular}
\end{table}

\section{Conclusions}
\label{sec:conclusions}

This research shows possibilities for developing text processing tools that can assist political scientists in developing new analytical tools. Computer scientists can help political scientists in upscaling their policy design research. This process would allow taking advantage of large quantities of legal documents on the Web.

Our prototype presented in the paper responds to three significant obstacles regarding building relevant data based on legal regulations: collecting relevant documents, extracting relevant information from many legal documents, and creating computer-readable data sets. It showcases that it is possible using wildly available tools to create a system that can collect large corpora of legal documents and generate graph representations of legal rules from them.

Modelling legal regulations through graphs using IG as an intermediate layer seems promising. (1) It is an opportunity of incorporating computer science technologies and methodologies into policy design studies. It makes it possible to produce comprehensive models of specific public policies relatively quickly, compare them and make quantitative analyses. (2) IG is a proper tool for capturing crucial elements of legal rules, which could be transformed into other data formats.  

Of course, our research is still developing, and we will be able to evaluate our system entirely only after finishing our work with other documents associated with the 2003 UNESCO Convention. It will be necessary to improve our tagger during this process to reduce manual annotation parts in applying IG for legal text annotation, e.g., splitting complex statements and identifying coreferences. 

The prototype presented in this paper also meets the challenge of using network analysis techniques to express legal rules in an automated way in a computer-readable format. This task is accomplished using IG as an intermediate layer between a legal text and mainly graph-based metrics expressing interrelations between entities. Therefore, the prototype system presented in this paper is constructed to allow users to extract computer-readable data efficiently from legal regulations. 

\section{Acknowledgments}
The research was funded by the Centre for Priority Research Area Artificial Intelligence and Robotics of Warsaw University of Technology within the Excellence Initiative: Research University (IDUB) programme (grant no 1820/27/Z01/POB2/2021).

Hanna Schreiber wishes to acknowledge that her contribution to this chapter was carried out within the framework of the research grant Sonata 15, "Between the heritage of the world and the heritage of humanity: researching international heritage regimes through the prism of Elinor Ostrom’s IAD framework", 2019/35/D/HS5/04247 financed by the National Science Centre (Poland).

We would like to thank many Students of Data Science in the Faculty of Mathematics and Information Science for their work under the guidance of Anna Wróblewska and Bartosz Pieliński on Institutional Grammar taggers and preliminary ideas of the system (e.g. \cite{wichrowska_system_2021}), which we modified and extended further.
\FloatBarrier

\bibliographystyle{unsrtnat}
\bibliography{references}  

\end{document}